\documentclass[11pt]{article}
\usepackage{acl2015}
\usepackage{times}
\usepackage{url}
\usepackage{latexsym}
\usepackage{comment}
\usepackage{amsmath,amsthm,amssymb}
\usepackage[vlined]{algorithm2e}
\usepackage{graphicx}
\usepackage{multirow}
\usepackage{color}
\usepackage[normalem]{ulem}
\usepackage{skt}
\usepackage[utf8]{inputenc}
\usepackage[russian,english]{babel}

\newcommand{\vect}[1]{\mathbf{#1}}

\newcommand{\ignore}[1]{}

\title{Non-distributional Word Vector Representations}

\author{Manaal Faruqui \and Chris Dyer\\
  Language Technologies Institute \\
  Carnegie Mellon University \\
  Pittsburgh, PA, 15213, USA \\
  {\tt \{mfaruqui, cdyer\}@cs.cmu.edu}
  }

\date{}

\begin{document}
\maketitle

\begin{abstract}
  Data-driven representation learning for words is a technique
  of central importance in NLP. While indisputably useful  as a source of
  features in downstream tasks, such vectors tend to consist of uninterpretable
  components whose relationship to the categories of traditional lexical
  semantic theories is tenuous at best. We
  present a method for constructing interpretable word vectors from
  hand-crafted linguistic resources like WordNet, FrameNet etc. These
  vectors are binary (i.e, contain only 0 and 1) and are 99.9\% sparse.
  We analyze their performance on state-of-the-art evaluation methods for
  distributional models of word vectors and find they are competitive to
standard distributional approaches.
\end{abstract}

  \ignore{
Distributional word vector representations are constructed using
word cooccurrence information from large unannotated corpora.
These methods of word vector training usually ignore the valuable
information about words present in widely available linguistic resources like
WordNet, FrameNet, etc. Further, the dimensions of word vectors trained using
these approaches are uninterpretable. In this paper we construct
word vectors solely using linguistic information about words
and show that these vectors are either better or competitive to
distributional word vectors trained on billions of words and
achieve state-of-the-art performance on a word similarity task. Since,
the dimensions of these word vectors are linguistic features they are
fully interpretable.
}

\section{Introduction}
\label{sec:intro}

Distributed representations of words have been shown to benefit a diverse set of NLP tasks including
syntactic parsing \cite{lazaridou:2013,bansal:2014}, named
entity recognition \cite{guo2014revisiting} and sentiment
analysis \cite{socher:2013}. Additionally, because they can be induced directly from unannotated corpora, they are likewise available in domains and languages where traditional linguistic resources do not exhaust.
Intrinsic evaluations on various tasks are helping refine vector learning methods to discover representations that captures many facts about lexical
semantics \cite{Turney:2001:MWS:645328.650004,Turney10fromfrequency}.

Yet induced word vectors do not look anything like the representations described in
most lexical semantic theories, which focus on identifying classes of words
\cite{verb-classes.levin.1993,baker:1998,Schuler:2005:VBC:1104493,miller:1995}.
Though expensive to construct,
conceptualizing word meanings symbolically is important
for theoretical understanding and interpretability is desired in computational
models.

Our contribution to this discussion is a new technique that constructs
task-independent word vector representations using linguistic knowledge derived
from pre-constructed linguistic resources like WordNet \cite{miller:1995},
FrameNet \cite{baker:1998}, Penn Treebank \cite{marcus1993building} etc.
In such word vectors every
dimension is a linguistic feature and 1/0 indicates the presence or absence of
that feature in a word, thus the vector representations are binary while being
highly sparse ($\approx 99.9\%$). Since these vectors do not encode any word
cooccurrence information, they are non-distributional.
An additional benefit of constructing such
vectors is that they are fully interpretable i.e, every dimension of these
vectors maps to a linguistic feature unlike distributional
word vectors where the vector dimensions have no interpretability.

Of course, engineering feature vectors from linguistic resources is established
practice in many applications of discriminative learning; e.g., parsing
\cite{mcdonald2006online,nivre2008algorithms} or part of
speech tagging \cite{ratnaparkhi1996maximum,collins2002discriminative}.
However, despite a certain common inventories of features that re-appear across
many tasks, feature engineering tends to be seen as a task-specific problem,
and engineered feature vectors are not typically evaluated independently of the
tasks they are designed for. We evaluate the quality of our linguistic
vectors on a number of tasks that have been proposed for evaluating
distributional word vectors. We show that linguistic word vectors are
comparable to current state-of-the-art distributional word
vectors trained on billions of words as evaluated on a battery of semantic and
syntactic evaluation benchmarks.\footnote{Our vectors can be downloaded at:
\url{https://github.com/mfaruqui/non-distributional}}

\section{Linguistic Word Vectors}

\begin{table}[!tb]
  \centering
  \small
  \begin{tabular}{|l|c|c|}
  \hline
  Lexicon & Vocabulary & Features \\
  \hline
  WordNet & 10,794 & 92,117 \\
  Supersense & 71,836 & 54\\
  FrameNet & 9,462 & 4,221\\
  Emotion & 6,468 & 10\\
  Connotation & 76,134 & 12\\
  Color & 14,182 & 12\\
  Part of Speech & 35,606 & 20\\
  Syn. \& Ant. & 35,693 & 75,972\\
  \hline
  Union & 119,257 & 172,418\\
  \hline
  \end{tabular}
    \caption{Sizes of vocabualry and features induced from different
    linguistic resources.}
  \label{tab:size}
\end{table}

\begin{table*}[!tb]
  \centering
  \small
  \begin{tabular}{|l|c|c|c|c|c|c|c|}
  \hline
  Word & \textsc{Pol.Pos} & \textsc{Color.Pink} & \textsc{SS.Noun.Feeling} & \textsc{PTB.Verb} & \textsc{Anto.Fair} & $\cdots$ & \textsc{Con.Noun.Pos} \\
  \hline
  love & 1 & 1 & 1 & 1 & 0 & & 1\\
  hate & 0 & 0 & 1 & 1 & 0 & & 0\\
  ugly & 0 & 0 & 0 & 0 & 1 & & 0\\
  beauty & 1 & 1 & 0 & 0 & 0 & & 1\\
  refundable & 0 & 0 & 0 & 0 & 0 & & 1\\

  \hline
  \end{tabular}
    \caption{Some linguistic word vectors. 1 indicates presence and 0 indicates absence of a linguistic feature.}
  \label{tab:ling}
\end{table*}

We construct linguistic word vectors by extracting word level information
from linguistic resources. Table~\ref{tab:size} shows the size of
vocabulary and number of features induced from every lexicon.
We now describe various linguistic resources that we use for
constructing linguistic word vectors.

\paragraph{WordNet.} WordNet \cite{miller:1995} is an English lexical database
that groups words
into sets of synonyms called synsets and records a number of relations among
these synsets or their members. For a word we look up its synset for all
possible part of speech (POS) tags that it can assume.
For example, \textit{film} will have \textsc{Synset.Film.V.01} and
\textsc{Synset.Film.N.01} as features as it can be both a verb and a noun.
In addition to synsets, we include the hyponym (for ex.
\textsc{Hypo.CollageFilm.N.01}), hypernym (for ex. \textsc{Hyper:Sheet.N.06})
and holonym synset of the word as features. We also collect antonyms and
pertainyms
of all the words in a synset and include those as features in the linguistic
vector. 

\paragraph{Supsersenses.} WordNet partitions nouns and verbs into semantic field
categories known as supsersenses \cite{Ciaramita:2006,nastase:2008}.
For example, \textit{lioness} evokes the supersense \textsc{SS.Noun.Animal}.
These supersenses were further extended to adjectives
\cite{tsvetkov:2014}.\footnote{\url{http://www.cs.cmu.edu/~ytsvetko/adj-supersenses.tar.gz}}
We use these supsersense tags for nouns, verbs and adjectives as features in the
linguistic word vectors. 

\paragraph{FrameNet.} FrameNet \cite{baker:1998,fillmore-ua-2003} is a rich
linguistic resource that
contains information about lexical and predicate-argument semantics in English.
Frames can be realized on the surface by many different word types,
which suggests that the word types evoking the same frame should be semantically
related. For every word, we use the frame it evokes along with the roles of the
evoked frame as its features. Since, information in FrameNet is
part of speech (POS) disambiguated, we couple these feature with the
corresponding POS tag of
the word. For example, since \textit{appreciate} is a verb, it will have
the following features: \textsc{Verb.Frame.Regard}, \textsc{Verb.Frame.Role.Evaluee} etc.

\paragraph{Emotion \& Sentiment.}
\newcite{mohammad:2013} constructed two different
lexicons that associate words to sentiment polarity and to emotions resp. using
crowdsourcing. The polarity is either positive or negative but there are eight
different kinds of emotions like anger, anticipation, joy etc. Every word in the
lexicon is associated with these properties.
For example, \textit{cannibal} evokes \textsc{Pol.Neg},
\textsc{Emo.Disgust} and \textsc{Emo.Fear}. We use these properties as
features in linguistic vectors. 

\paragraph{Connotation.} \newcite{feng:2013} construct a lexicon that contains
information about connotation of words that are seemingly objective but often
allude nuanced sentiment. They assign positive, negative and neutral
connotations to these words. This lexicon differs from
\newcite{mohammad:2013} in
that it has a more subtle shade of sentiment and it extends to many more words.
For example, \textit{delay} has a negative connotation \textsc{Con.Noun.Neg}, \textit{floral} has a positive connotation \textsc{Con.Adj.Pos} and
\textit{outline} has a neutral connotation \textsc{Con.Verb.Neut}.

\paragraph{Color.} Most languages have expressions involving color, for example
\textit{green with envy} and \textit{grey with uncertainly} are phrases used in
English. The word-color associtation lexicon produced by
\newcite{mohammad:2011}
using crowdsourcing lists the colors that a word
evokes in English. We use every color in this lexicon as a feature in the
vector. For example, \textsc{Color.Red} is a feature evoked by the word
\textit{blood}.

\paragraph{Part of Speech Tags.} The Penn Treebank \cite{marcus1993building}
annotates naturally
occurring text for linguistic structure. It contains syntactic parse trees and
POS tags for every word in the corpus. We collect all the possible POS tags
that a word is annotated with and use it as features in the linguistic vectors.
For example, \textit{love} has \textsc{PTB.Noun}, \textsc{PTB.Verb} as features.

\paragraph{Synonymy \& Antonymy.} We use Roget's thesaurus \cite{Roget11} to
collect sets of synonymous words.\footnote{\url{http://www.gutenberg.org/ebooks/10681}}
For every word, its synonymous word is used
as a feature in the linguistic vector. For example, \textit{adoration} and
\textit{affair} have a feature \textsc{Syno.Love}, \textit{admissible} has
a feature \textsc{Syno.Acceptable}. The synonym lexicon contains 25,338 words
after removal
of multiword phrases. In a similar manner, we also use antonymy
relations between words as features in the word vector. The antonymous words
for a given word were collected from
\newcite{ordway1913synonyms}.\footnote{\url{https://archive.org/details/synonymsantonyms00ordwiala}}
An example
would be of \textit{impartiality}, which has features \textsc{Anto.Favoritism}
and \textsc{Anto.Injustice}. The antonym lexicon has 10,355 words. These
features are different from those induced from WordNet as the former encode
word-word relations whereas the latter encode word-synset relations.

After collecting features from the various linguistic resources described above
we obtain linguistic word vectors of length 172,418 dimensions. These vectors
are 99.9\% sparse i.e, each vector on an average
contains only 34 non-zero features out of 172,418 total features.
On average a linguistic feature (vector dimension) is active for 15 word types.
The linguistic word vectors contain 119,257 unique word types.
Table~\ref{tab:ling} shows linguistic vectors for
some of the words.

\section{Experiments}

We first briefly describe the evaluation tasks and then present
results.

\begin{table*}[!tb]
  \centering
  \small
  \begin{tabular}{|l|l|l|l||c|c|c||c|c|}
  \hline
  Vector & Length ($D$) & Params. & Corpus Size & WS-353 & RG-65 & SimLex & Senti & NP \\
  \hline
  Skip-Gram & 300 & $D\times N$& 300 billion & 65.6 & 72.8	& 43.6 & \textbf{81.5}	& 80.1\\
  Glove & 300 & $D\times N$ &  6 billion & 60.5 & 76.6  & 36.9 & 77.7 & 77.9\\
  LSA & 300 & $D\times N$ & 1 billion & \textbf{67.3} & 77.0 & 49.6 & 81.1 & 79.7 \\
  \hline
  Ling Sparse & 172,418 & -- & -- & 44.6 & \textbf{77.8}  & 56.6 & 79.4 & \textbf{83.3}\\
  Ling Dense & 300 & $D\times N$ & -- & 45.4 & 67.0  & \textbf{57.8} & 75.4 & 76.2\\
  \hline\hline
  Skip-Gram $\oplus$ Ling Sparse & 172,718 & -- & -- & 67.1 & 80.5 & 55.5 & 82.4 & 82.8\\
  \hline
  \end{tabular}
    \caption{Performance of different type of word vectors on evaluation tasks
    reported by Spearman's correlation (first 3 columns) and Accuracy (last 2
    columns). Bold shows the best performance for a task.} 
  \label{tab:eval}
\end{table*}

\subsection{Evaluation Tasks}

\paragraph{Word Similarity.}
We evaluate our word representations on three different benchmarks to measure
word similarity. The first one is the widely used WS-353 dataset
\cite{citeulike:379845}, which contains 353 pairs of English words that have
been assigned similarity ratings by humans. The second is the RG-65 dataset
\cite{Rubenstein:1965} of 65 words pairs. The third dataset is SimLex
\cite{HillRK14} which has been constructed to overcome the shortcomings of
WS-353 and contains 999 pairs of adjectives, nouns and verbs.
Word similarity is computed using cosine similarity between two words and
Spearman's rank correlation is reported
between the rankings produced by vector model against the human rankings.

\paragraph{Sentiment Analysis.}
\newcite{socher:2013} created a treebank containing sentences
annotated with fine-grained sentiment labels on phrases and sentences
from movie review excerpts.
The coarse-grained treebank of positive and negative
classes has been split into training, development, and test datasets
containing 6,920, 872, and 1,821 sentences, respectively.
We use average of the word vectors of a given sentence as features in
an $\ell_2$-regularized logistic regression for classification. The classifier
is tuned on the dev set and accuracy is reported on the test set.

\paragraph{NP-Bracketing.} \newcite{lazaridou:2013}
constructed a dataset from the Penn TreeBank \cite{marcus1993building} of
noun phrases (NP) of length three words, where the first can be an adjective or
a noun and the other two are nouns. The task is to predict the correct
bracketing in the parse tree for a given noun phrase. For example,
\textit{local (phone company)} and \textit{(blood pressure) medicine} exhibit
\textit{left} and \textit{right} bracketing respectively.
We append the word vectors
of the three words in the NP in order and use them as features in an
$\ell_2$-regularized logistic regression classifier.
The dataset contains 2,227 noun phrases split into 10 folds.
The classifier is tuned on the first fold and cross-validation accuracy
is reported on the remaining nine folds.

\subsection{Linguistic Vs. Distributional Vectors}

In order to
make our linguistic vectors comparable to publicly available distributional
word vectors, we perform singular value decompostion (SVD) on
the linguistic matrix to obtain word vectors of lower dimensionality. If
$\vect{L} \in \{0,1\}^{N\times D}$ is the linguistic matrix with $N$ word types
and $D$ linguistic features, then we can obtain
$\vect{U} \in \mathbb{R}^{N\times K}$ from the SVD of
$\vect{L}$ as follows: $\vect{L} = \vect{U}\vect{\Sigma}\vect{V}^{\top}$, with
$K$ being the desired length of the lower dimensional space.

We compare both sparse and dense linguistic vectors to three widely used
distributional word vector models. The first two are the pre-trained Skip-Gram
\cite{mikolov:2013}\footnote{\url{https://code.google.com/p/word2vec}} and
Glove
\cite{glove:2014}\footnote{\url{http://www-nlp.stanford.edu/projects/glove/}}
word vectors each of length 300, trained on 300 billion and 6 billion words
respectively. We used latent semantic analysis (LSA) to obtain word vectors from
the SVD decomposition of a word-word cooccurrence matrix
\cite{Turney10fromfrequency}. These were trained on 1 billion words of
Wikipedia with vector length 300 and context window of 5 words.

\subsection{Results}

Table~\ref{tab:eval} shows the performance of different word vector types on the
evaluation tasks. It can be seen that although Skip-Gram, Glove \& LSA perform
better than
linguistic vectors on WS-353, the linguistic vectors outperform them
by a huge margin on SimLex.
Linguistic vectors also perform better at RG-65.
On sentiment analysis, linguistic vectors are competitive with Skip-Gram vectors
and on the NP-bracketing task they outperform all distributional vectors with a
statistically significant margin (p $<$ 0.05, McNemar's test
\newcite{Dietterich98}). We append the sparse linguistic vectors to Skip-Gram
vectors and evaluate the resultant vectors as shown in the bottom row of Table
\ref{tab:eval}. The combined vector outperforms Skip-Gram on all tasks,
showing that linguistic vectors contain useful information orthogonal to
distributional information.

It is evident from the results that linguistic vectors are
either competitive or better to state-of-the-art distributional vector models.
Sparse linguistic word vectors are high dimensional but they are
also sparse, which makes them computationally easy to work with.

\section{Discussion}

Linguistic resources like WordNet have found extensive applications in lexical
semantics, for example, for word sense disambiguation, word similarity etc.
\cite{Resnik:1995,agirre:2009}.
Recently there has been interest in using linguistic resources to enrich word
vector representations. In these approaches, relational information among words
obtained from WordNet, Freebase etc. is used as a constraint to encourage words
with similar properties in lexical ontologies to have similar word vectors
\cite{xu:2014,Yu:2014,yan-ecml14,fried:2014,faruqui:2015}. Distributional
representations have also been shown to improve by using experiential data in
addition to distributional context \cite{andrews2009integrating}.
We have shown that
simple vector concatenation can likewise be used to improve representations
(further confirming the established finding that lexical resources and
cooccurrence information provide somewhat orthogonal information), but it is
certain that more careful combination strategies can be used.

Although distributional word vector dimensions cannot, in general, be
identified with linguistic  properties, it has been shown that some vector
construction strategies yield dimensions that are relatively more interpretable
\cite{murphy:2012,fyshe:2014,fyshe:2015,faruqui:2015:sparse}. However, such
analysis is difficult to generalize across models of representation. In
constrast to distributional word vectors, linguistic word vectors have
interpretable dimensions as every dimension is a linguistic property.

Linguistic word vectors require no training as there are no parameters to be
optimized, meaning they are computationally
economical. While good quality
linguistic word vectors may only be obtained for languages with rich
linguistic resources, such resources do exist in many languages and should not
be disregarded.

\section{Conclusion}

We have presented a novel method of constructing word vector representations
solely using linguistic knowledge from pre-existing linguistic resources.
These non-distributional, linguistic word vectors are competitive
to the current models of distributional word vectors as evaluated
on a battery of tasks. Linguistic vectors are fully interpretable as every
dimension is a linguistic feature and are highly sparse, so they are
computationally easy to work with.

\section*{Acknowledgement}
We thank Nathan Schneider for giving comments on an earlier draft
of this paper and the anonymous reviewers for their feedback.

\bibliographystyle{acl}
\bibliography{references}

\begin{thebibliography}{}

\bibitem[\protect\citename{Agirre \bgroup et al.\egroup }2009]{agirre:2009}
Eneko Agirre, Enrique Alfonseca, Keith Hall, Jana Kravalova, Marius Pa\c{s}ca,
  and Aitor Soroa.
\newblock 2009.
\newblock A study on similarity and relatedness using distributional and
  wordnet-based approaches.
\newblock In {\em Proc. of NAACL}.

\bibitem[\protect\citename{Andrews \bgroup et al.\egroup
  }2009]{andrews2009integrating}
Mark Andrews, Gabriella Vigliocco, and David Vinson.
\newblock 2009.
\newblock Integrating experiential and distributional data to learn semantic
  representations.
\newblock {\em Psychological review}, 116(3):463.

\bibitem[\protect\citename{Baker \bgroup et al.\egroup }1998]{baker:1998}
Collin~F. Baker, Charles~J. Fillmore, and John~B. Lowe.
\newblock 1998.
\newblock The berkeley framenet project.
\newblock In {\em Proc. of ACL}.

\bibitem[\protect\citename{Bansal \bgroup et al.\egroup }2014]{bansal:2014}
Mohit Bansal, Kevin Gimpel, and Karen Livescu.
\newblock 2014.
\newblock Tailoring continuous word representations for dependency parsing.
\newblock In {\em Proc. of ACL}.

\bibitem[\protect\citename{Bian \bgroup et al.\egroup }2014]{yan-ecml14}
Jiang Bian, Bin Gao, and Tie-Yan Liu.
\newblock 2014.
\newblock Knowledge-powered deep learning for word embedding.
\newblock In {\em Proc. of MLKDD}.

\bibitem[\protect\citename{Ciaramita and Altun}2006]{Ciaramita:2006}
Massimiliano Ciaramita and Yasemin Altun.
\newblock 2006.
\newblock Broad-coverage sense disambiguation and information extraction with a
  supersense sequence tagger.
\newblock In {\em Proc. of EMNLP}.

\bibitem[\protect\citename{Collins}2002]{collins2002discriminative}
Michael Collins.
\newblock 2002.
\newblock Discriminative training methods for hidden markov models: Theory and
  experiments with perceptron algorithms.
\newblock In {\em Proc. of EMNLP}.

\bibitem[\protect\citename{Dietterich}1998]{Dietterich98}
Thomas~G. Dietterich.
\newblock 1998.
\newblock Approximate statistical tests for comparing supervised classification
  learning algorithms.
\newblock {\em Neural Computation}.

\bibitem[\protect\citename{Faruqui \bgroup et al.\egroup }2015a]{faruqui:2015}
Manaal Faruqui, Jesse Dodge, Sujay~K. Jauhar, Chris Dyer, Eduard Hovy, and
  Noah~A. Smith.
\newblock 2015a.
\newblock Retrofitting word vectors to semantic lexicons.
\newblock In {\em Proc. of NAACL}.

\bibitem[\protect\citename{Faruqui \bgroup et al.\egroup
  }2015b]{faruqui:2015:sparse}
Manaal Faruqui, Yulia Tsvetkov, Dani Yogatama, Chris Dyer, and Noah~A. Smith.
\newblock 2015b.
\newblock Sparse overcomplete word vector representations.
\newblock In {\em Proc. of ACL}.

\bibitem[\protect\citename{Feng \bgroup et al.\egroup }2013]{feng:2013}
Song Feng, Jun~Seok Kang, Polina Kuznetsova, and Yejin Choi.
\newblock 2013.
\newblock Connotation lexicon: A dash of sentiment beneath the surface meaning.
\newblock In {\em Proc. of ACL}.

\bibitem[\protect\citename{Fillmore \bgroup et al.\egroup
  }2003]{fillmore-ua-2003}
Charles Fillmore, Christopher Johnson, and Miriam Petruck.
\newblock 2003.
\newblock Lexicographic relevance: selecting information from corpus evidence.
\newblock {\em International Journal of Lexicography}.

\bibitem[\protect\citename{Finkelstein \bgroup et al.\egroup
  }2001]{citeulike:379845}
Lev Finkelstein, Evgeniy Gabrilovich, Yossi Matias, Ehud Rivlin, Zach Solan,
  Gadi Wolfman, and Eytan Ruppin.
\newblock 2001.
\newblock {Placing search in context: the concept revisited}.
\newblock In {\em Proc. of WWW}.

\bibitem[\protect\citename{Fried and Duh}2014]{fried:2014}
Daniel Fried and Kevin Duh.
\newblock 2014.
\newblock Incorporating both distributional and relational semantics in word
  representations.
\newblock {\em arXiv preprint arXiv:1412.4369}.

\bibitem[\protect\citename{Fyshe \bgroup et al.\egroup }2014]{fyshe:2014}
Alona Fyshe, Partha~P. Talukdar, Brian Murphy, and Tom~M. Mitchell.
\newblock 2014.
\newblock Interpretable semantic vectors from a joint model of brain- and text-
  based meaning.
\newblock In {\em Proc. of ACL}.

\bibitem[\protect\citename{Fyshe \bgroup et al.\egroup }2015]{fyshe:2015}
Alona Fyshe, Leila Wehbe, Partha~P. Talukdar, Brian Murphy, and Tom~M.
  Mitchell.
\newblock 2015.
\newblock A compositional and interpretable semantic space.
\newblock In {\em Proc. of NAACL}.

\bibitem[\protect\citename{Guo \bgroup et al.\egroup }2014]{guo2014revisiting}
Jiang Guo, Wanxiang Che, Haifeng Wang, and Ting Liu.
\newblock 2014.
\newblock Revisiting embedding features for simple semi-supervised learning.
\newblock In {\em Proc. of EMNLP}.

\bibitem[\protect\citename{Hill \bgroup et al.\egroup }2014]{HillRK14}
Felix Hill, Roi Reichart, and Anna Korhonen.
\newblock 2014.
\newblock Simlex-999: Evaluating semantic models with (genuine) similarity
  estimation.
\newblock {\em CoRR}, abs/1408.3456.

\bibitem[\protect\citename{Lazaridou \bgroup et al.\egroup
  }2013]{lazaridou:2013}
Angeliki Lazaridou, Eva~Maria Vecchi, and Marco Baroni.
\newblock 2013.
\newblock Fish transporters and miracle homes: How compositional distributional
  semantics can help {NP} parsing.
\newblock In {\em Proc. of EMNLP}.

\bibitem[\protect\citename{Levin}1993]{verb-classes.levin.1993}
Beth Levin.
\newblock 1993.
\newblock {\em English verb classes and alternations : a preliminary
  investigation}.
\newblock University of Chicago Press.

\bibitem[\protect\citename{Marcus \bgroup et al.\egroup
  }1993]{marcus1993building}
Mitchell~P Marcus, Mary~Ann Marcinkiewicz, and Beatrice Santorini.
\newblock 1993.
\newblock Building a large annotated corpus of english: The penn treebank.
\newblock {\em Computational linguistics}, 19(2):313--330.

\bibitem[\protect\citename{McDonald and Pereira}2006]{mcdonald2006online}
Ryan~T McDonald and Fernando~CN Pereira.
\newblock 2006.
\newblock Online learning of approximate dependency parsing algorithms.
\newblock In {\em Proc. of EACL}.

\bibitem[\protect\citename{Mikolov \bgroup et al.\egroup }2013]{mikolov:2013}
Tomas Mikolov, Kai Chen, Greg Corrado, and Jeffrey Dean.
\newblock 2013.
\newblock Efficient estimation of word representations in vector space.
\newblock {\em arXiv preprint arXiv:1301.3781}.

\bibitem[\protect\citename{Miller}1995]{miller:1995}
George~A Miller.
\newblock 1995.
\newblock Wordnet: a lexical database for english.
\newblock {\em Communications of the ACM}.

\bibitem[\protect\citename{Mohammad and Turney}2013]{mohammad:2013}
Saif~M. Mohammad and Peter~D. Turney.
\newblock 2013.
\newblock Crowdsourcing a word-emotion association lexicon.
\newblock {\em Computational Intelligence}, 29(3):436--465.

\bibitem[\protect\citename{Mohammad}2011]{mohammad:2011}
Saif Mohammad.
\newblock 2011.
\newblock Colourful language: Measuring word-colour associations.
\newblock In {\em Proc. of the Workshop on Cognitive Modeling and Computational
  Linguistics}.

\bibitem[\protect\citename{Murphy \bgroup et al.\egroup }2012]{murphy:2012}
Brian Murphy, Partha Talukdar, and Tom Mitchell.
\newblock 2012.
\newblock Learning effective and interpretable semantic models using
  non-negative sparse embedding.
\newblock In {\em Proc. of COLING}.

\bibitem[\protect\citename{Nastase}2008]{nastase:2008}
Vivi Nastase.
\newblock 2008.
\newblock Unsupervised all-words word sense disambiguation with grammatical
  dependencies.
\newblock In {\em Proc. of IJCNLP}.

\bibitem[\protect\citename{Nivre}2008]{nivre2008algorithms}
Joakim Nivre.
\newblock 2008.
\newblock Algorithms for deterministic incremental dependency parsing.
\newblock {\em Computational Linguistics}, 34(4):513--553.

\bibitem[\protect\citename{Ordway}1913]{ordway1913synonyms}
Edith~Bertha Ordway.
\newblock 1913.
\newblock {\em Synonyms and Antonyms: An Alphabetical List of Words in Common
  Use, Grouped with Others of Similar and Opposite Meaning}.
\newblock Sully and Kleinteich.

\bibitem[\protect\citename{Pennington \bgroup et al.\egroup }2014]{glove:2014}
Jeffrey Pennington, Richard Socher, and Christopher~D. Manning.
\newblock 2014.
\newblock Glove: Global vectors for word representation.
\newblock In {\em Proc. of EMNLP}.

\bibitem[\protect\citename{Ratnaparkhi}1996]{ratnaparkhi1996maximum}
Adwait Ratnaparkhi.
\newblock 1996.
\newblock A maximum entropy model for part-of-speech tagging.
\newblock In {\em Proc. of EMNLP}.

\bibitem[\protect\citename{Resnik}1995]{Resnik:1995}
Philip Resnik.
\newblock 1995.
\newblock Using information content to evaluate semantic similarity in a
  taxonomy.
\newblock In {\em Proc. of IJCAI}.

\bibitem[\protect\citename{Roget}1852]{Roget11}
P.~M. Roget.
\newblock 1852.
\newblock {\em {Roget's Thesaurus of English words and phrases}}.
\newblock Available from Project Gutemberg.

\bibitem[\protect\citename{Rubenstein and Goodenough}1965]{Rubenstein:1965}
Herbert Rubenstein and John~B. Goodenough.
\newblock 1965.
\newblock Contextual correlates of synonymy.
\newblock {\em Commun. ACM}, 8(10).

\bibitem[\protect\citename{Schuler}2005]{Schuler:2005:VBC:1104493}
Karin~Kipper Schuler.
\newblock 2005.
\newblock {\em Verbnet: A Broad-coverage, Comprehensive Verb Lexicon}.
\newblock {Ph.D.} thesis, University of Pennsylvania.

\bibitem[\protect\citename{Socher \bgroup et al.\egroup }2013]{socher:2013}
Richard Socher, Alex Perelygin, Jean Wu, Jason Chuang, Christopher~D. Manning,
  Andrew~Y. Ng, and Christopher Potts.
\newblock 2013.
\newblock Recursive deep models for semantic compositionality over a sentiment
  treebank.
\newblock In {\em Proc. of EMNLP}.

\bibitem[\protect\citename{Tsvetkov \bgroup et al.\egroup }2014]{tsvetkov:2014}
Yulia Tsvetkov, Nathan Schneider, Dirk Hovy, Archna Bhatia, Manaal Faruqui, and
  Chris Dyer.
\newblock 2014.
\newblock Augmenting english adjective senses with supersenses.
\newblock In {\em Proc. of LREC}.

\bibitem[\protect\citename{Turney and Pantel}2010]{Turney10fromfrequency}
Peter~D. Turney and Patrick Pantel.
\newblock 2010.
\newblock From frequency to meaning : Vector space models of semantics.
\newblock {\em JAIR}, pages 141--188.

\bibitem[\protect\citename{Turney}2001]{Turney:2001:MWS:645328.650004}
Peter~D. Turney.
\newblock 2001.
\newblock Mining the web for synonyms: Pmi-ir versus lsa on toefl.
\newblock In {\em Proc. of ECML}.

\bibitem[\protect\citename{Xu \bgroup et al.\egroup }2014]{xu:2014}
Chang Xu, Yalong Bai, Jiang Bian, Bin Gao, Gang Wang, Xiaoguang Liu, and
  Tie-Yan Liu.
\newblock 2014.
\newblock Rc-net: A general framework for incorporating knowledge into word
  representations.
\newblock In {\em Proc. of CIKM}.

\bibitem[\protect\citename{Yu and Dredze}2014]{Yu:2014}
Mo~Yu and Mark Dredze.
\newblock 2014.
\newblock Improving lexical embeddings with semantic knowledge.
\newblock In {\em Proc. of ACL}.

\end{thebibliography}
\end{document}